# Suivi temps réel du SoH de batteries lithium-ion

Bruno JAMMES, Edgar Hernando SEPULVEDA-OVIEDO, Corinne ALONSO
LAAS-CNRS, Université de Toulouse, CNRS, Toulouse, France

**RESUME**

Le suivi en temps réel de l'état de santé (SoH) des batteries reste un défi majeur, particulièrement dans les micro-réseaux où les contraintes opérationnelles limitent l'usage des méthodes traditionnelles. Dans le cadre du projet 4BLife, nous proposons une méthode innovante basée sur l'analyse d'une impulsion de décharge en fin de charge. Les paramètres du modèle électrique équivalent décrivant l'évolution de la tension aux bornes de batterie sur cette impulsion de courant sont ensuite utilisés pour estimer le SoH. Sur la base de données expérimentales acquises à ce jour, les premiers résultats montrent la pertinence de l'approche proposée. Après un apprentissage utilisant les paramètres de deux batteries présentant une dégradation de la capacité de l'ordre de 85%, nous avons réussi à prévoir la dégradation de deux autres batteries, cyclées jusqu'à environ 90% de SoH, avec un écart absolu moyen de l'ordre de 1% dans le pire des cas, et une explicabilité de l'estimateur voisine de 0,9.

Si les performances de cette méthode se confirment, elle peut être facilement intégrée dans les systèmes de gestion des batteries (BMS) et ouvre la voie à une gestion optimisée des batteries en exploitation continue.

*Mots-clés—État de santé (SoH), Modèle électrique équivalent, Micro-réseaux, Gestion des batteries (BMS), Cyclage des batteries, Dégradation des batteries.*

## 1. INTRODUCTION

Le suivi en fonctionnement de l'état de santé (SoH) des batteries reste un défi majeur dans les systèmes de gestion des micro-réseaux. Les systèmes de gestion des batteries (BMS) disponibles intègrent souvent des estimateurs de vieillissement qui, selon les retours d'expérience des utilisateurs, manquent encore de fiabilité et de précision. Ce constat est appuyé par plusieurs travaux [1], qui soulignent l'importance d'intégrer des modèles précis de vieillissement pour garantir une gestion optimale des micro-réseaux.

D'autre part, les méthodes conventionnelles d'estimation du SoH nécessitant généralement une décharge complète des batteries se révèlent inadaptées pour des batteries en exploitation continue. Dans les applications mobiles, où la disponibilité de l'énergie est également critique, ces limitations deviennent encore plus problématiques, car une estimation fiable du SoH est essentielle pour éviter des interruptions inattendues et maximiser l'autonomie.

Plusieurs études dans la littérature ont exploré des méthodes avancées pour l'estimation du SoH, notamment des approches basées sur des modèles physiques et des algorithmes avancés comme [2]. Dans le cadre du projet 4BLife, nous avons exploré des solutions innovantes pour une estimation fiable du SoH en temps réel, basées sur l'analyse des paramètres électriques durant une brève impulsion de décharge et dans un état de charge contrôlé.

## 2. METHODE

Nous proposons ici une estimation du SoH basée sur les paramètres du modèle électrique équivalent décrivant l'évolution de la tension aux bornes de la batterie. Cette approche a déjà été utilisée dans la littérature, où différentes méthodes exploitent les paramètres issus de modèles électriques équivalents (ECM) pour prédire le SoH à partir de phases de relaxation [3], de cycles complets optimisés [4], ou de données embarquées enrichies par des caractéristiques IC (capacité incrémentale) et thermiques [5].

L'originalité de notre proposition réside dans l'analyse de la réponse en tension à une impulsion de décharge appliquée en fin de charge, permettant l'identification de paramètres dynamiques corrélés à l'état de santé. Ces paramètres sont ensuite exploités par un estimateur entraîné sur un jeu de batteries, puis appliqués à d'autres cellules de même technologie. La figure 1 présente un schéma synthétique de la méthodologie mise en œuvre, illustrant les différentes étapes du processus allant de l'application de l'impulsion de courant jusqu'à l'estimation du SoH en temps réel.

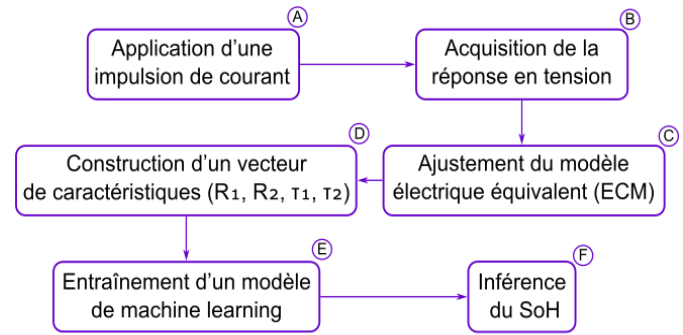

Figure 1 : Schéma de la méthodologie proposée pour l'estimation du SoH basée sur l'analyse d'une impulsion de décharge en fin de charge.

## 3. EXPERIMENTATION

L'étude expérimentale, lancée en avril 2023, dans le cadre du projet 4BLIFE (c.f §9), est toujours en cours. Elle est menée sur quatre batteries lithium-ion LiFeP04 (105Ah - 3.2V) fournies par la société BATConnect et soumises à des cycles de charge-décharge à température maîtrisée (21°C). Ce protocole a permis de collecter à ce jour 4 jeux des paramètres du modèle électrique dans des conditions contrôlées et sur des SoH variant de 100% à 85%. L'évolution de ces SoH pour les quatre batteries testées est présentée à la figure 2.

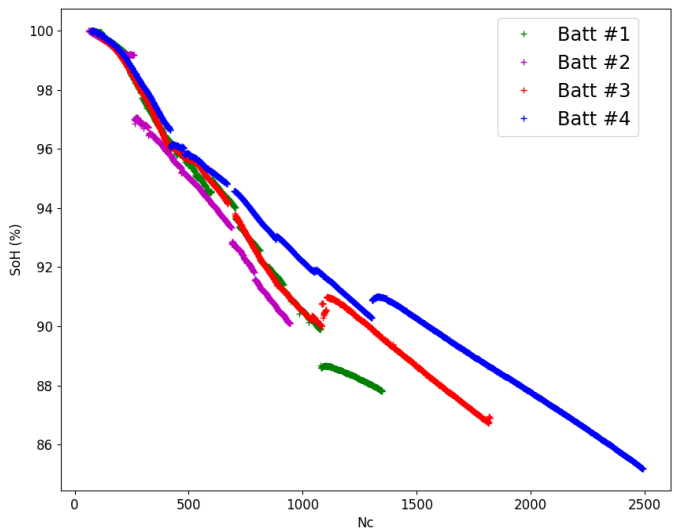

Figure 2 : Évolution de l'état de santé (SoH) en fonction du nombre de cycles pour les 4 batteries testées.

## 4. MODELISATION DE L'IMPULSION DE DECHARGE

L'évolution de la tension de la batterie étant analysée sur une très courte durée, nous pouvons considérer que son état de charge ne varie pas pendant la mesure. La variation de la tension aux bornes des batteries est alors modélisée par :

$$\Delta E_{cell} = I_{cell} * \left(R_{int} + R_1 * \left(1 - e^{-\frac{t}{\tau_1}}\right) + R_2 * \left(1 - e^{-\frac{t}{\tau_2}}\right)\right)$$

La comparaison de la réponse du modèle à des données expérimentales (cf. figure 3) montre une oscillation persistante du modèle autour des données qui laisse penser qu'il y a en réalité plus de 2 constantes de temps dans la réponse réelle.

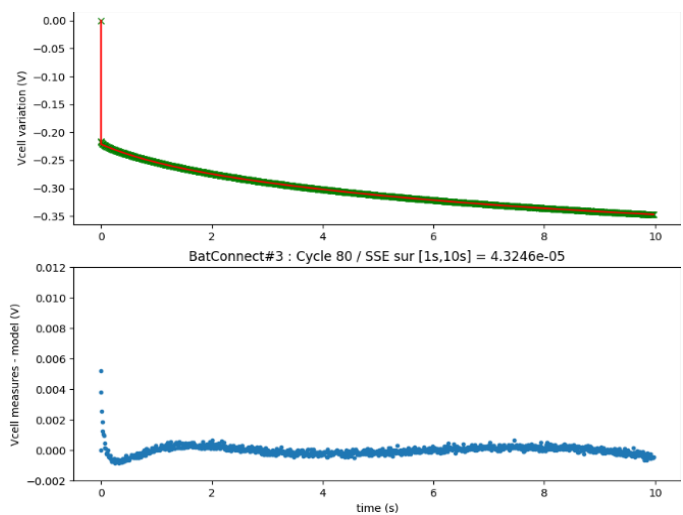

Figure 3 : Modélisation de la tension batterie sur 10s pour un échelon de courant de -60A. Courbe du haut = comparaison données/modèle, courbe du bas = erreur de modélisation.

L'augmentation de l'ordre du modèle n'a toutefois pas conduit à une réduction significative de ces oscillations. Par contre, la qualité globale du modèle, quantifiée par la somme des erreurs quadratiques (SSE) sur l'intervalle [1s, 10s] s'améliore significativement en ne considérant pas la 1ère seconde des données expérimentales lors de l'identification des paramètres (cf. figure 4).

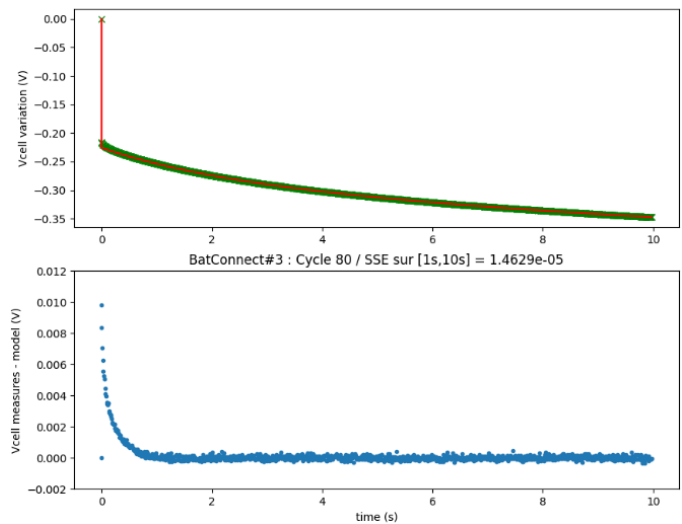

Figure 4 : Modélisation de la tension batterie sur 10s pour un échelon de courant de -60A en utilisant uniquement les données de 1 à 10s. Courbe du haut = comparaison données/modèle, courbe du bas = erreur de modélisation.

Suite à cette modification de la procédure d'identification des paramètres du modèle, nous pouvons noter une réduction de la dispersion des paramètres tout au long de l'expérimentation. Afin de garantir une mesure fiable in situ de la constante de temps, plusieurs précautions sont requises. L'impulsion de courant est appliquée en fin de charge, lorsque l'état de charge (SoC) est stable, ce qui permet de supposer des conditions initiales homogènes. La température ambiante est contrôlée à 21 °C, et les mesures sont réalisées à courant constant, assurant la reproductibilité du profil de décharge. La constante de temps $\tau_1$ est obtenue par ajustement des paramètres du modèle électrique équivalent à la réponse en tension mesurée sur une fenêtre de 1 à 10 secondes après l'impulsion. Ce processus peut être facilement automatisé dans un système de gestion de batterie (BMS), ce qui rend possible une estimation périodique du SoH en fonctionnement réel.

## 5. EVOLUTION DES PARAMETRES AVEC LE VIEILLISSEMENT

L'analyse des paramètres du modèle après plusieurs centaines de cycles nous a conforté dans notre démarche car les paramètres du modèle semblent bien corrélés au SoH. On peut voir sur la figure 5 que le SoH est bien corrélé aux paramètres des éléments dynamiques du modèle, pour les 4 batteries testées. Les points présentés sur la figure 5 représentent la moyenne des paramètres sur une fenêtre glissante de 20 échantillons, l'évolution du SoH étant considérée peu significative sur 20 cycles consécutifs. Un des intérêts de la méthode proposée qui peut être appliquée à la fin de chaque recharge est de permettre un filtrage très simple des paramètres.

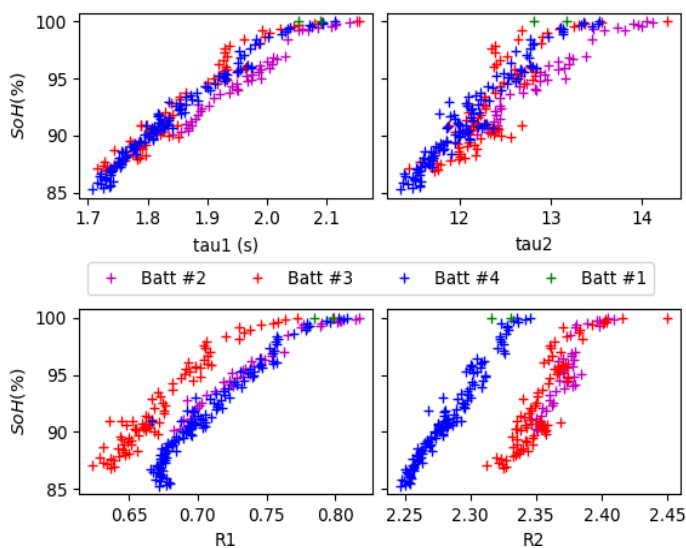

Figure 5 : Évolution du SoH en fonction des 4 paramètres du modèle moyennés sur 20 cycles, pour les 4 batteries testées.

A contrario, les valeurs de la résistance interne ($R_{int}$) et leurs évolutions en fonction du SoH s'avèrent très différentes suivant les batteries (cf. figure 6).

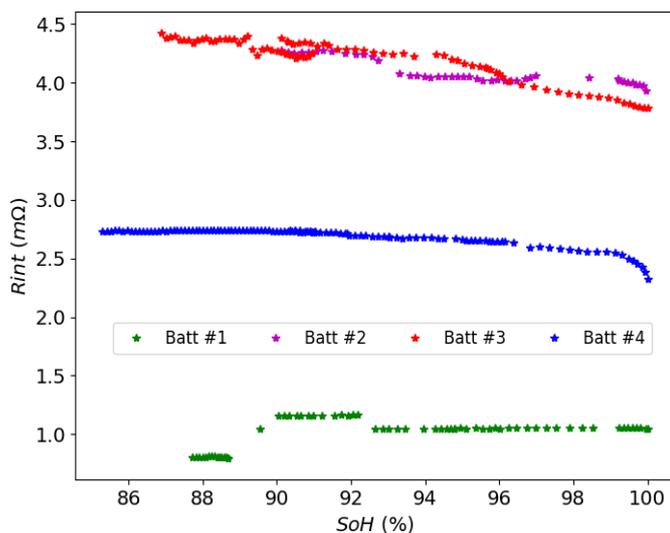

Figure 6 : Évolution de la résistance interne estimée moyennée sur 2 cycles en fonction du SoH pour les 4 batteries testées.

De plus, la mesure de cette résistance est, dans notre cas, sensible au montage/démontage des batteries, en particulier pour la batterie 1. Ce constat nous a récemment conduits à revoir le câblage de l'expérimentation. Toutefois, dans cette étude nous ne prendrons pas en compte cette résistance. Comme indiqué précédemment, toutes les grandeurs utilisées ont été moyennées sur une fenêtre glissante de 20 échantillons.

6. APPRENTISSAGE DE L'ESTIMATEUR DE VIEILLISSEMENT

Le vecteur d'entrées utilisé pour l'apprentissage automatique est composé des paramètres $R_1$, $R_2$, $\tau_1$ et $\tau_2$. Dans cette étude nous avons écarté la phase de « rodage » des batteries qui étaient neuves au début des essais. Ainsi, pour chaque batterie, nous considérons les cycles à partir du celui pour lequel la capacité déchargée est maximale. Ceci nous a conduit à écarter entre les 60 à 80 premiers cycles suivant la batterie.

Plusieurs modèles de régression ont été testés afin d'estimer le SoH à partir des paramètres sélectionnés comme des modèles classiques (régression linéaire, Support Vector Regression, K-plus-proches-voisins, arbres de décision), des apprentissages d'ensembles (Random Forest, Gradient Boosting, Extreme Gradient Boosting, ...), des apprentissages robustes (Theil-Sen, Huber), des réseaux de neurones multi-couches et un modèle de deep learning implémenté avec la bibliothèque Python Keras.

Nous avons commencé par entraîner l'estimateur de SoH sur les données des batteries 3 et 4 qui présentent la plus grande variation de SoH. Ce protocole permet d'évaluer la robustesse de l'estimateur sur deux batteries tout en évitant le surapprentissage spécifique à une seule batterie. Des premiers entrainements, il ressort que ce sont les valeurs fournies par la régression linéaire qui se rapprochent le plus des données expérimentales pour les 2 batteries de test (cf. figures 7 et 8).

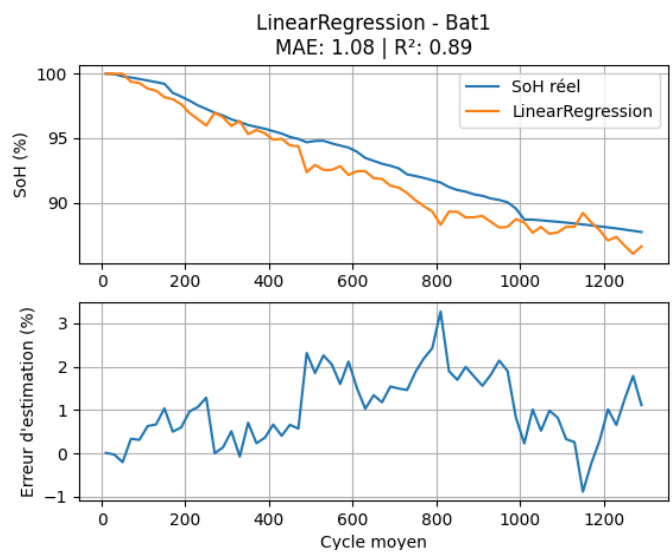

Figure 7 : Prédiction du SoH pour la batterie 1 à l'aide d'un modèle linéaire entraîné sur les batteries 3 et 4.

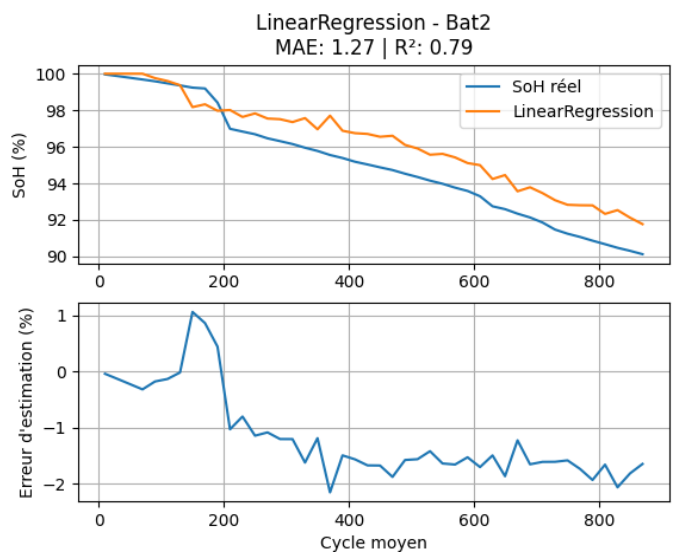

Figure 8 : Prédiction du SoH pour la batterie 2 à l'aide d'un modèle linéaire entraîné sur les batteries 3 et 4.

Nous avons utilisé 2 indicateurs pour qualifier les performances de l'estimateur : la moyenne des erreurs absolues (MAE) et le coefficient de détermination $R^2$. Ce dernier reflète la proportion de la variance du SoH expliquée par le modèle.

Le bon comportement de la régression linéaire (MAE de l'ordre de 1.27% et $R^2$ de l'ordre de 0.8) pourrait s'expliquer par une relation quasi-affine entre certains paramètres du modèle (notamment $\tau_1$ et $\tau_2$) et le SoH. Cette démarche simple offre ainsi un bon compromis entre une estimation du SoH relativement précise et la simplicité d'implémentation, ce qui justifie son usage dans un contexte embarqué.

Toutefois, la taille modeste du jeu de données ne couvrant à ce jour qu'un horizon de SoH limité et la faible variabilité des conditions expérimentales en température et courant de charge/décharge, notamment en température, limitent la capacité des modèles d'apprentissage profond. Cela explique en partie pourquoi ces méthodes ne surpassent pas les approches plus simples dans notre cas. Il est également probable que l'utilisation des valeurs par défaut des hyperparamètres n'ait pas permis aux modèles non linéaires d'exprimer leur plein potentiel. Une optimisation via des techniques comme Optuna ou la recherche bayésienne pourrait améliorer leurs performances, tout en limitant le surapprentissage.

Le comportement de l'estimateur sur la batterie 2 a retenu notre attention car il est globalement bien au-dessus de la réalité. Comme on peut le voir sur la figure 9 (courbe magenta), le vieillissement de cette batterie présente une forte discontinuité, sans raison apparente, aux alentours du $50^{ième}$ cycle.

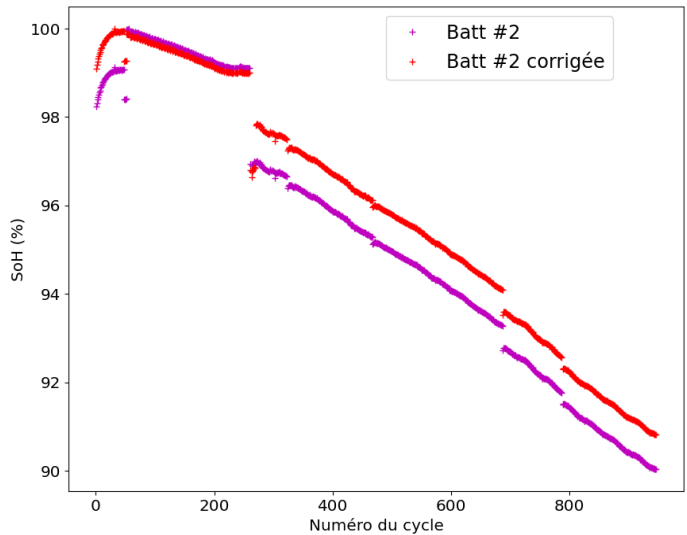

Figure 9 : A.h déchargés originaux pour la batterie 2 sur la totalité des tests (magenta) et proposition de recalage (rouge).

Nous proposons donc d'enlever 1A.h à la capacité récupérée sur les cycles 51 à 259. Nous n'avons pas corrigé les autres décalages car ils correspondent à des arrêts de l'expérimentation avec perte des données (cependant, cela ne permet pas d'expliquer entièrement la perte de capacité importante constatée à la reprise des tests) et/ou arrêt du test pendant quelques jours dans des conditions particulièrement stressantes pour la batterie (SoC = 0).

Après correction du SoH, la prévision du SoH pour la batterie 2 se rapproche sensiblement de l'expérimentation (figure 10) : la MAE passant de 1,27% à 0,58% et le $R^2$ de 0,79 à 0,94.

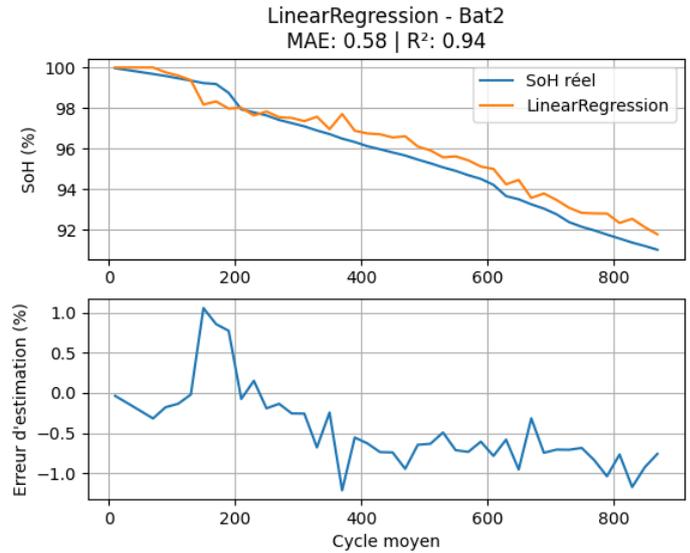

Figure 20 : Comparaison du SoH corrigé de la batterie 2 avec le modèle linéaire entraîné sur les batteries 3 et 4.

## 7. DISCUSSION

Les résultats que nous venons de présenter nous confortent dans la possibilité d'estimer le SoH d'une batterie à partir de simples paramètres d'un modèle équivalent électrique d'ordre 2 décrivant la réponse en tension à une impulsion de courant de 10s. Parmi les méthodes d'apprentissage testées, le modèle linéaire est un de ceux qui donne à ce jour les meilleures performances : une moyenne des erreurs absolues de l'ordre de 1% dans le pire des cas et un pouvoir explicatif de l'ordre de 0,9.

La constante de temps $\tau_1$, significativement corrélée au SoH dans nos expérimentations, est interprétée dans cette étude comme reflétant principalement des dynamiques électrochimiques rapides à l'interface électrode-électrolyte. Bien que cette interprétation ne soit pas directement prouvée ici, elle s'appuie sur des travaux antérieurs ayant établi des relations entre les paramètres de modèles électriques équivalents et les phénomènes électrochimiques internes, tels que la croissance de la couche solide électrolytique interphase (SEI) ou les résistances interfaciales [6]. Selon cette hypothèse, le vieillissement des cellules ralentirait ces processus rapides, entraînant une augmentation progressive de $\tau_1$. Cette tendance ferait ainsi de $\tau_1$ un indicateur sensible et pertinent pour le suivi de la dégradation de l'état de santé.

La simplicité de la méthode, combinée à son faible impact sur le fonctionnement des batteries, représente une avancée importante par rapport aux approches conventionnelles qui nécessitent des calculs complexes (modèles physiques détaillés) ou des configurations expérimentales exigeantes (décharges complètes), comme décrit dans [7].

Cependant, certaines limites de cette approche doivent être précisées : le cyclage a été réalisé dans des conditions de température maîtrisée, avec des décharges complètes, un intervalle de SoH restreint (100 à 85%) et sur un petit nombre de batteries d'une même technologie. Il est donc nécessaire, pour valider définitivement la méthode proposée, de poursuivre les tests, puis de faire varier la température et de mettre en œuvre des décharges partielles par exemple.

Même dans le cas où l'obtention de données pour apprendre l'estimateur de SoH ne serait pas envisageable, cette méthode pourrait être envisagée pour comparer le vieillissement entre cellules ou packs dans une installation en fonctionnement, sans avoir de modélisation préalable à effectuer, uniquement en comparant l'évolution des paramètres du modèle électrique équivalent, et en considérant éventuellement le SoH estimé par le BMS. Les techniques d'intelligence artificielle testées à ce jour n'ont pas fait mieux que la régression linéaire dans notre cas mais ayant déjà démontré leur efficacité dans l'extraction de signatures caractéristiques [8], la prédiction précise [9] ou le diagnostic avancé [10] de l'état de santé des systèmes photovoltaïques ; devraient trouver alors de l'intérêt.

8. CONCLUSIONS

La méthode proposée dans le cadre du projet 4BLife offre une approche simple, novatrice et efficace pour le suivi de l'état de santé (SoH) des batteries LiFeP04 en s'appuyant sur l'analyse d'impulsions de décharge de faible durée dans des conditions bien identifiées (fin de charge). Contrairement aux méthodes conventionnelles nécessitant une décharge complète, cette approche permet une estimation en temps réel sans interrompre le fonctionnement des batteries. Cette méthode, simple et non invasive, représente une solution prometteuse pour améliorer la gestion des batteries en exploitation continue. Les premiers résultats, sur une base de données pour l'instant modeste, montrent qu'un estimateur de SoH appris sur les 2 batteries qui ont le plus vieilli présente des performances tout à fait honorables sur les deux autres batteries que nous avons cyclées, la batterie 1 étant celle qui s'éloigne le plus du modèle avec une moyenne des erreurs absolues de l'ordre du 1.08% et une erreur absolue globalement en dessous de 2% pouvant exceptionnellement monter jusqu'à 3%.

Les expériences se poursuivent pour valider l'approche, en élargissant les tests à différentes conditions environnementales. Nous travaillons également du côté des paramètres des algorithmes d'apprentissage pour améliorer leurs performances en termes de généralisation.